\documentclass[letterpaper,10pt,conference]{ieeeconf}

\usepackage[T1]{fontenc}
\usepackage[utf8]{inputenc}
\usepackage{times}
\usepackage{microtype}
\usepackage{cite}
\usepackage{graphicx}
\usepackage{booktabs}
\usepackage{flushend}
\usepackage{amsmath,amssymb}
\usepackage{array}
\usepackage{xcolor}
\usepackage{url}
\usepackage{tikz}
\usetikzlibrary{arrows.meta,positioning,fit}

\IEEEoverridecommandlockouts
\definecolor{terrain}{RGB}{42,119,83}
\definecolor{motion}{RGB}{40,99,155}
\definecolor{contact}{RGB}{190,100,37}

\title{\LARGE \bf
From Target Selection to Digging: A Learning-Based Framework for Continuous Autonomous Excavation
}

\author{Shuai Zhao$^{1}$, Ji-an Pan$^{2,*}$, Quantao Yang$^{3}$, Zheng Wang$^{2}$,\\
Chaoyi Chen$^{4}$, Qing Xu$^{4}$, and Keqiang Li$^{4}$%
\thanks{$^{1}$Faculty of Information, Liaoning University.}%
\thanks{$^{2}$School of Mechanical Engineering and Automation, Northeastern University.}%
\thanks{$^{3}$Department of Robotics, Perception, and Learning, KTH Royal Institute of Technology, Sweden.}%
\thanks{$^{4}$School of Vehicle and Mobility, Tsinghua University.}%
\thanks{$^{*}$Corresponding author: Ji-an Pan.}%
}

\begin{document}
\maketitle
\thispagestyle{empty}
\pagestyle{empty}

\begin{abstract}
Repeated excavation continuously reshapes pile geometry, requiring an autonomous excavator to adapt its digging targets and coordinate motion across successive excavation cycles. We present a learning-based framework for continuous autonomous excavation that integrates terrain-aware target selection with reinforcement- and imitation-learning controllers. The framework separates target-conditioned motion from local digging: a shared task-conditioned RL policy controls waypoint-guided approach and loaded transport, while an IL policy learns vision-based digging and lifting from expert demonstrations. Digging targets are selected from LiDAR elevation maps and converted into bucket-tip waypoints for motion control. The control architecture coordinates the learned policies and deterministic unloading through a shared motion interface. The complete system is deployed on a scaled hydraulic excavator with multimodal sensing and closed-loop actuator control. Offline replay and physical experiments demonstrate more consistent target selection, shorter local motion time, and increased payload compared with the respective baselines. The learned digging policy achieves a mean payload of 6.52 kg per completed cycle, compared with 2.68 kg for Fixed Dig. Three five-scoop runs further demonstrate consecutive autonomous excavation under continuously changing pile geometry.

\end{abstract}

\section{Introduction}
\label{sec:introduction}

Excavation and bulk material handling are fundamental operations in mining, construction, and port logistics~\cite{spinelli2026material}. Every scoop reshapes the source pile, requiring the operator to continually select new digging locations and coordinate approach, digging, transport, and unloading. These repetitive operations depend on skilled operators and are often performed under demanding conditions. An integrated autonomous excavation system that combines terrain perception, task planning, and hydraulic control could reduce operator workload and support consistent material transfer.

\begin{figure}[!t]
\centering
\includegraphics[width=\columnwidth]{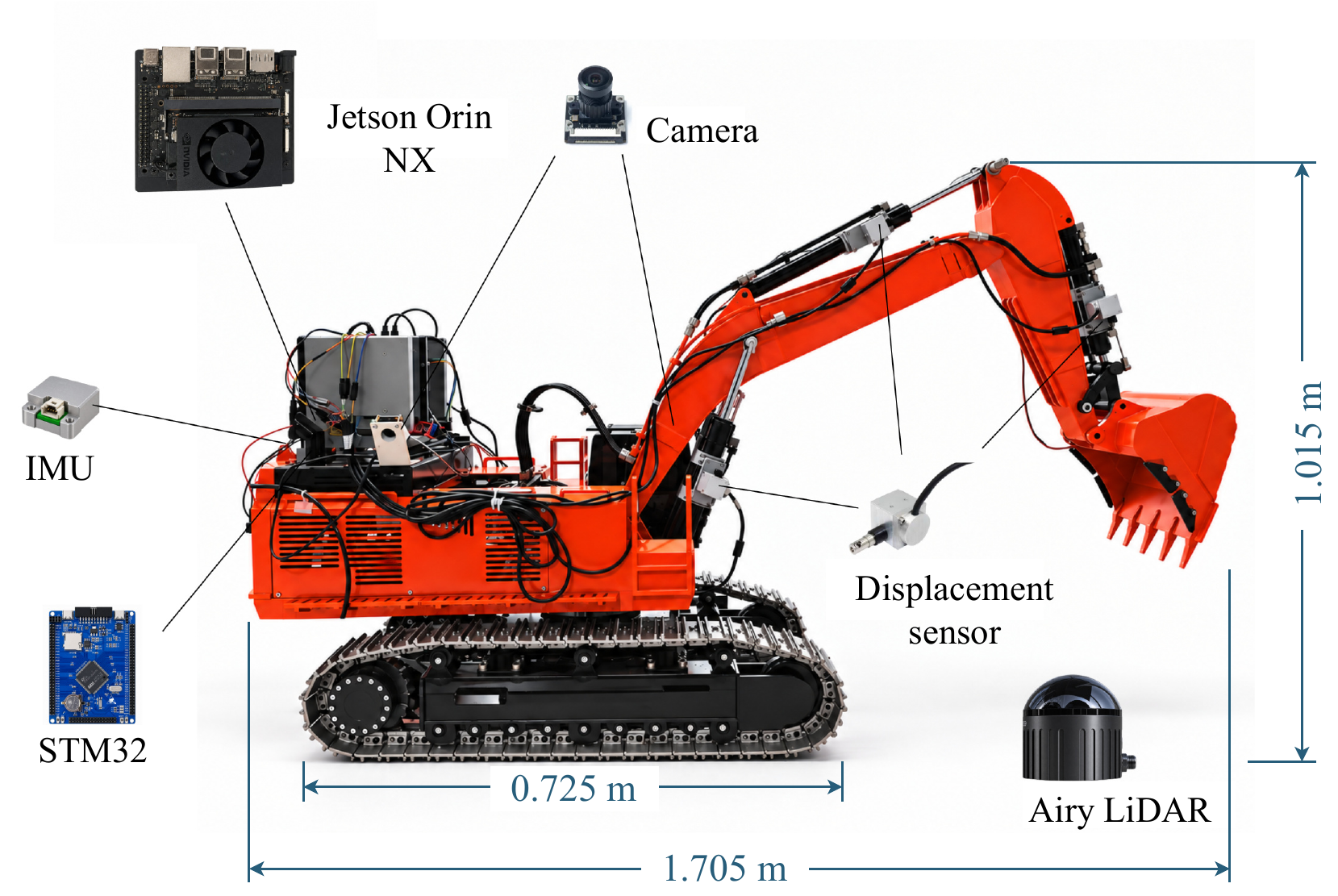}
\caption{Hardware Platform and System Integration. We robotize a teleoperated scaled hydraulic excavator by integrating multimodal sensing, actuator feedback, Jetson Orin NX policy inference, and STM32-based low-level control. Dimensions show overall length and height in the photographed pose.}
\label{fig:system}
\end{figure}

Trajectory optimization~\cite{yang2021optimization}, expert-guided trajectory generation~\cite{guo2022imitation,shen2024hybrid}, and online predictive control~\cite{okada2024database} have enabled excavation motions to account for terrain geometry, operator experience, and changing interaction conditions. Integrated systems have advanced repeated excavation~\cite{jud2017planning,zhang2021autonomous,terenzi2023towards}, earth cutting with onboard inspection~\cite{jang2025integrated}, and learned target selection with motion control for material handling~\cite{spinelli2026material}. Despite this progress, connecting continuously changing terrain targets to learning-based excavation remains challenging because target-conditioned motion and material acquisition impose different control objectives and supervision requirements.

Target-conditioned motions, such as approach and loaded transport, can be specified by bucket-tip waypoints, but require different bucket orientations when reaching the pile and retaining material during transport. Digging, in contrast, requires coordinated penetration, curling, and lifting that can be learned directly from expert demonstrations. This distinction suggests separating target-conditioned motion from local digging skill, while coordinating both within a common excavation cycle. Meanwhile, hydraulic dead zones complicate accurate execution of learned motion, and observation noise can cause unstable target updates as the pile evolves. Together, these challenges motivate a system that jointly addresses terrain-aware target selection, target-conditioned motion, and demonstration-trained digging for continuous autonomous excavation.

We propose a learning-based framework for continuous autonomous excavation that combines terrain-aware target selection, target-conditioned motion, and demonstration-trained digging. The Temporal Target Selector updates digging targets from LiDAR elevation maps as the pile evolves, while the RL Tracker uses a shared, task-conditioned RL waypoint-tracking policy for empty-bucket approach and loaded transport. For local digging and lifting, we train an ACT-based imitation-learning (IL) policy~\cite{zhao2023act} on real expert demonstrations. The control architecture coordinates these components with deterministic unloading and repeated terrain observation, forming a closed excavation cycle that adapts to the evolving pile geometry. Our hardware platform and system integration enable deployment on a scaled hydraulic excavator (Fig.~\ref{fig:system}) through LiDAR and visual sensing, actuator feedback, onboard policy inference, and low-level closed-loop control. Our contributions are:
\begin{enumerate}
\item A closed-loop learning-based framework for continuous autonomous excavation that separates target-conditioned, waypoint-guided motion from local digging. Repeated terrain observation and terrain-aware target selection update the digging goal after each scoop.

\item A control architecture coordinating one task-conditioned RL policy for empty-bucket approach and loaded transport, demonstration-trained ACT for digging and lifting, and Fixed Dump through a shared motion interface. Hardware platform and system integration enable multimodal sensing, actuator feedback, onboard inference, and closed-loop actuator control on a scaled hydraulic excavator.

\item Component-level and integrated real-machine evidence covering target selection, motion, digging, and continuous operation. Temporal selection reduces target jumps; ACT achieves 6.52~kg per completed cycle versus 2.68~kg for Fixed Dig; and three five-scoop runs complete without human intervention while sustaining material acquisition.

\end{enumerate}

\section{Related Work}
\label{sec:related}

\subsection{Autonomous Excavation Systems and Target Selection}

Autonomous excavation systems integrate terrain perception, planning, and control for successive cuts~\cite{jud2017planning,terenzi2023towards}, material loading~\cite{zhang2021autonomous}, and free-form trenching~\cite{jud2019trenching,jud2021heap}. Jang et al.~\cite{jang2025integrated} combine onboard landscape estimation, motion generation, and postexcavation inspection for earth cutting. Terrain maps also support constrained trajectory optimization~\cite{yang2021optimization}, while Guo et al.~\cite{guo2022imitation} use a highest-point rule to locate successive cuts. Spinelli et al.~\cite{spinelli2026material} combine learned LiDAR-based attack-point selection with path planning and RL motion control. Our target selection addresses temporal consistency across repeated terrain observations: the Temporal Target Selector filters and scores elevation-map candidates, then applies cross-frame confirmation and hysteresis before passing the target to a separate waypoint planner.

\subsection{Motion Control for Hydraulic Machines}

Hydraulic tracking methods address nonlinear and load-dependent actuator responses through predictive control~\cite{song2022lpv,okada2024database} or learned policies~\cite{egli2020towards,egli2022general,mori2025human}. Database-driven MPC updates the interaction model and reference trajectory during excavation~\cite{okada2024database}. Egli and Hutter~\cite{egli2022general} learn actuator dynamics from machine measurements and train an end-effector tracking policy in simulation. Spinelli et al.~\cite{spinelli2026material} use a waypoint-following policy to suppress gripper oscillations during approach and a separate policy for throwing. The RL Tracker instead shares one task-conditioned RL policy between empty-bucket approach and loaded transport. Its reward combines task-dependent bucket orientation objectives with penalties for commands within direction-dependent dead zones.

\subsection{Learning Excavation Skills}

Expert experience has been used to generate excavation trajectories~\cite{son2020expert, yang2025s}, initialize trajectory optimization~\cite{guo2022imitation}, and adapt motion primitives to new geometry~\cite{shen2024hybrid}. Action policies have also been learned for soil excavation and bucket filling~\cite{egli2022soil,egli2024bucket,dadhich2019field}, rigid-object excavation~\cite{jin2023rigid}, and boulder excavation~\cite{gruetter2025boulder}. These approaches differ in whether learning produces a reference trajectory or controls actions during material acquisition.

ACT predicts action chunks from current observations~\cite{zhao2023act}. ExACT~\cite{chen2024exact} learns reaching and digging--dumping sequences from real excavator demonstrations using RGB, joint positions, and task-dependent elevation-map inputs. Its evaluation uses recorded observations and a data-derived motion simulator. ExT~\cite{zhai2025ext} pretrains multitask policies from mixed expert demonstrations and demonstrates a complete excavation workflow on a real machine. ACT has also been evaluated on a small-scale physical construction platform~\cite{rasul2026physical}. Our distinction is the allocation of learning and observations across the cycle: LiDAR-based target selection and a shared RL motion policy handle changes in digging location, while standard ACT uses RGB and proprioception for local digging and lifting. Bucket-tip waypoint references connect the updated targets to this demonstration-trained skill. This design is evaluated through separate selection, motion, and digging comparisons and integrated five-scoop operation.

\section{Autonomous Excavation Framework}
\label{sec:formulation}
\label{sec:method}

\begin{figure*}[!t]
\centering
\includegraphics[width=\textwidth]{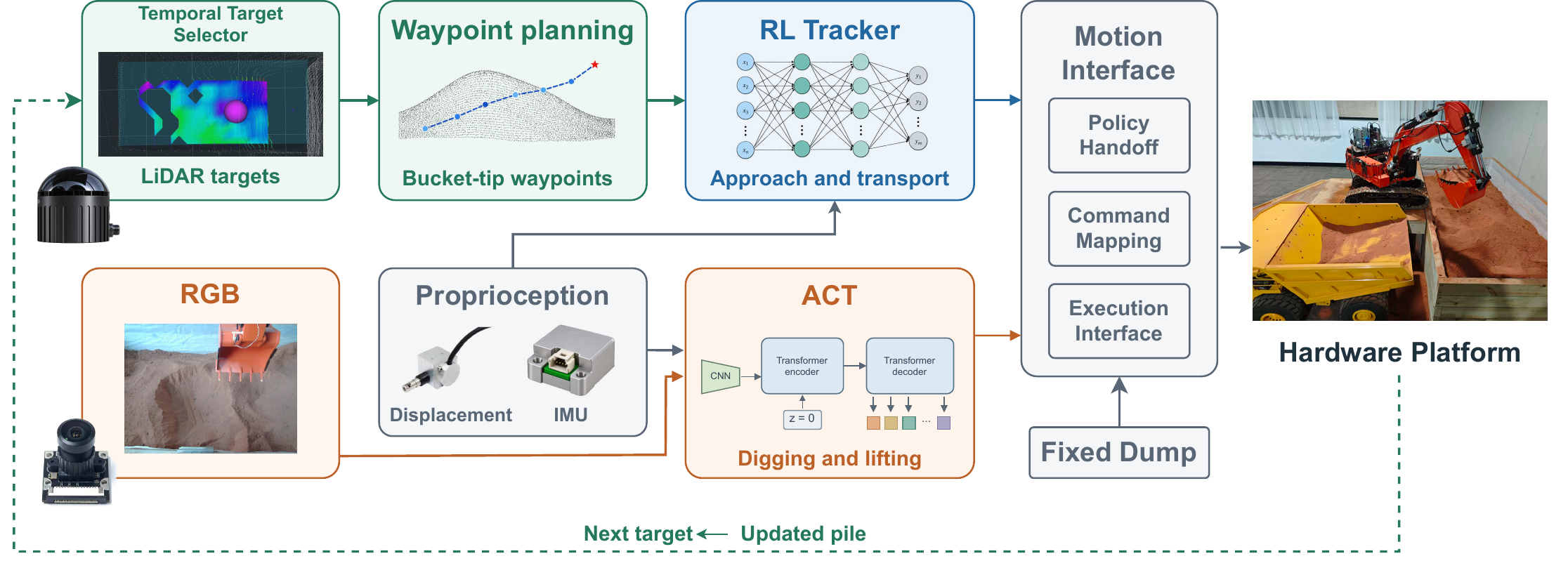}
\caption{Control Architecture. The Temporal Target Selector supplies targets for waypoint planning and shared RL control of approach and loaded transport. ACT uses RGB and proprioception for digging and lifting. Both policies and Fixed Dump share a motion interface. Solid arrows show data flow; the dashed arrow closes the cycle through renewed terrain observation.}
\label{fig:method}
\end{figure*}

Our framework (Fig.~\ref{fig:method}) combines three components: the Temporal Target Selector selects digging locations from LiDAR elevation maps; the RL Tracker uses a shared, task-conditioned RL policy for approach and loaded transport; and an ACT policy performs local digging and lifting from RGB and proprioception. A waypoint planner links terrain targets to motion, while a mission scheduler and shared motion interface coordinate the learned policies and deterministic unloading.

The design separates motion toward changing targets from local material acquisition. A pre-dig position and bucket pitch connect waypoint following to the demonstration-trained digging skill, which acts on the resulting image and machine state without an explicit terrain-goal input. After unloading, renewed terrain observation provides the target for the next scoop, closing the excavation cycle.

\subsection{Temporal Target Selector}

The Temporal Target Selector combines spatial suitability with temporal consistency. From ground and surface heights $H_g,H_s$, soil thickness is $H_d=\max(0,H_s-H_g)$. Cells meeting soil-height, confidence, and point-support requirements form a pile mask. Morphological opening and closing remove isolated noise and fill small holes; connected components identify candidate regions, and relative height, slope, boundary clearance, and local observation support further restrict eligible cells.

Each candidate $x$ receives a score
\begin{equation}
\begin{aligned}
S(x)={}&w_h\bar h(x)+w_c\bar c(x)+w_e e(x)\\
&+w_{\mathrm{rel}}r_h(x)-w_r\bar r(x),
\end{aligned}
\label{eq:tadps}
\end{equation}
where the nonnegative weights balance normalized local soil thickness $\bar h$, local confidence $\bar c$, boundary clearance $e$, preference for the admissible relative-height interval's center $r_h$, and roughness $\bar r$. A local surface fit supplies the selected target's height; waypoint planning handles reachability and path feasibility separately.

Candidates are associated across frames and confirmed before acceptance. A replacement must exceed the tracked score by a margin and pass temporal confirmation, while brief observation dropouts retain the current target. This hysteresis reduces observation-induced switches while allowing a different location to be selected as the pile changes. Settings appear in Section~\ref{sec:tadps_setup}.

\subsection{RL Tracker}

For scoop $i$, the selected target $g_i\in\mathbb{R}^3$ is converted into a pre-dig goal using an offset and workspace limits. A separate planner generates bucket-tip waypoints $\mathcal T_i=\{w_{i,k}\}_{k=0}^{K_i}$ toward either this goal or the unloading location. The RL Tracker shares one policy $\pi_F$ between these approach and transport motions, conditioned on task mode and target pitch. The pitch objective changes from pile entry to material retention, allowing the same motion policy to serve both stages.

The observation $o_t^F$ includes proprioception, bucket-tip state, waypoint errors, path progress, task mode, and pitch information, together with episode progress and the previous action. The policy outputs normalized cylinder-speed and swing-rate references. PPO~\cite{schulman2017ppo} trains the policy on randomized waypoint trajectories in simulation with reward
\begin{equation}
\begin{aligned}
r_t={}&w_p\Delta\rho_t-w_g e_t-w_c^{\mathrm{path}}c_t-w_\theta e_t^\theta\\
&-r_t^{\mathrm{act}}-r_t^{\mathrm{dz}}-r_t^{\mathrm{stop}}+b_t.
\end{aligned}
\label{eq:reward}
\end{equation}
Here, $\Delta\rho_t$ is waypoint-progress gain, and $e_t$, $c_t$, and $e_t^\theta$ penalize normalized waypoint distance, path-tube deviation, and absolute shortest-angle pitch error. The command cost $r_t^{\mathrm{act}}$ discourages abrupt or large actions; $r_t^{\mathrm{dz}}$ penalizes low-response nonzero commands within direction-dependent dead zones while preserving zero for stopping. While tracking the final loaded-transport waypoint, $r_t^{\mathrm{stop}}$ penalizes swing motion near the destination. The bonus $b_t$ rewards waypoint arrival and path completion. Settings appear in Section~\ref{sec:motion_setup}.

For each actuator and direction, let $D$ denote the dead-zone threshold with a margin. The dead-zone cost is proportional to $4z(1-z)$ for $0<|a|<D$, where $z=|a|/D$, and is zero otherwise. 

\subsection{Demonstration-Based Digging and Lifting}

Local digging requires coordinated penetration, curling, and lifting to acquire material. We learn this skill from real teleoperation using standard Action Chunking with Transformers (ACT)~\cite{zhao2023act}. Given front RGB $I_t$ and proprioception $p_t$, including cylinder, joint, and swing states, the policy predicts a sequence of normalized joystick commands:
\begin{equation}
\widehat{\mathbf A}_t=
\big(\hat{\mathbf a}_{t|t},\ldots,\hat{\mathbf a}_{t+H-1|t}\big)
=\pi_D(I_t,p_t).
\label{eq:act_chunk}
\end{equation}
Only the first $L<H$ commands are executed before querying the latest image and proprioception, unless the stage terminates earlier. This partial-chunk execution introduces observation feedback between action segments. Training data and chunk settings are reported in Section~\ref{sec:act_setup}.

\subsection{Control Architecture}
\label{sec:control_architecture}

A mission scheduler selects the active command source according to stage $q_t$:
\begin{equation}
\mathbf a_t=
\begin{cases}
\pi_F(o_t^F), & q_t\in\{\mathrm{move},\mathrm{carry}\},\\
\mathbf a_t^D, & q_t=\mathrm{dig},\\
\pi_{\mathrm{dump}}(p_t), & q_t=\mathrm{dump},
\end{cases}
\label{eq:hybrid}
\end{equation}
where $\mathbf a_t^D$ is the current command from the ACT chunk. Approach completion triggers ACT; ACT termination initiates loaded transport; and transport completion triggers Fixed Dump, followed by renewed terrain observation.

All policies use axis order [boom, stick, bucket, swing]. The shared motion interface preserves RL Tracker's velocity-reference and ACT's joystick semantics. A single motion controller transfers command authority between stages; both command paths share direction mapping, actuator limits, and a command watchdog. The hardware realization is described in Section~\ref{sec:platform_setup}.

\section{Experimental Setup}
\label{sec:experiments}

We evaluate target-selection stability, motion execution and transfer, digging payload, and consecutive autonomous operation. Component studies examine each module's behavior; integrated runs test their coordination over repeated cycles. Unless stated otherwise, $\pm$ denotes sample standard deviation.

\subsection{Hardware Platform and System Integration}
\label{sec:platform_setup}

Starting from a teleoperated scaled hydraulic excavator (Fig.~\ref{fig:system}), we integrate sensing, command and telemetry interfaces, onboard policy inference, and closed-loop actuator control to enable autonomous excavation. Boom, stick, and bucket actuation are hydraulic; swing is electric. Cable-displacement encoders measure cylinder positions; velocities are estimated from these signals. Boom and bucket angles use voltage-to-angle mappings, stick angle follows linkage geometry, and an IMU supplies yaw and rate. The central tooth-tip position is estimated through URDF forward kinematics.

LiDAR provides elevation maps and a 5-cm-voxel OctoMap; front RGB and timestamped proprioception supply policy observations. A host PC handles perception and planning, Jetson Orin NX runs the policies, and STM32 drives the actuators. Velocity control combines direction-specific static feedforward with proportional feedback. Positive/negative speed limits from displacement--time records are 35.1/18.5, 44.4/35.7, and 34.2/41.9~mm/s for boom, stick, and bucket. Both policies use common final actuator limits.

All physical experiments use one soil type with the machine stationary. Payload is measured using a 500-kg scale with 0.05-kg display division. The truck rests on the scale and is emptied and re-tared before each measurement. Component comparisons weigh individual scoops; continuous runs weigh five-scoop totals, with per-scoop payloads additionally recorded.

\begin{figure*}[t]
\centering
\includegraphics[width=0.90\textwidth]{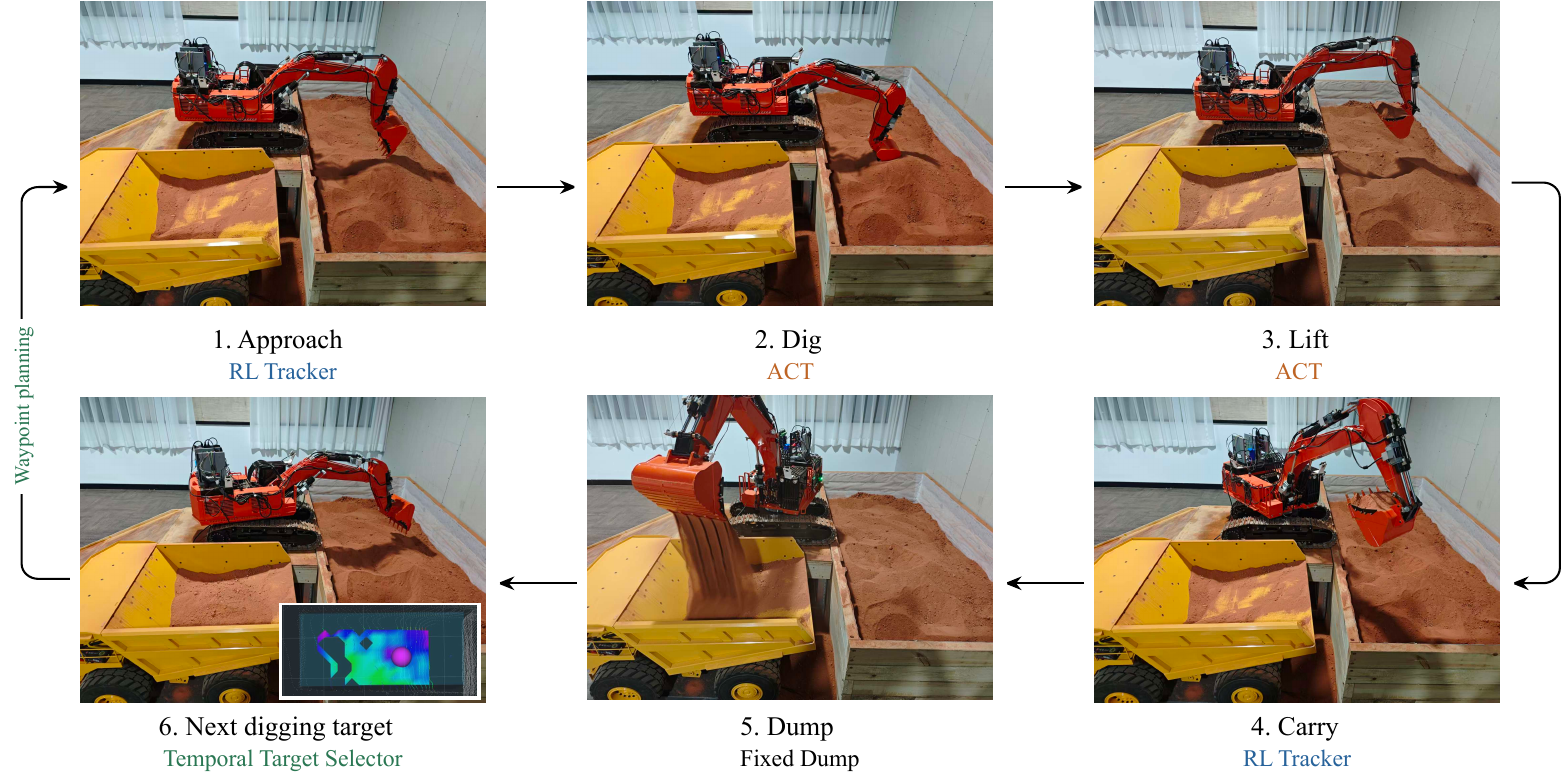}
\caption{Repeated excavation workflow. The RL Tracker handles approach and loaded transport; ACT performs digging and lifting; Fixed Dump unloads. The Temporal Target Selector selects the next target, followed by waypoint planning before the next approach. The inset in panel 6 shows a captured visualization of the Temporal Target Selector with the selected target marked by a magenta sphere.}
\label{fig:continuous_photos}
\end{figure*}

\subsection{Implementation and Training}
\label{sec:implementation}

\subsubsection{Target selection}
\label{sec:tadps_setup}
The Temporal Target Selector uses 10-cm map cells. The weights in Eq.~\eqref{eq:tadps} are 0.35 for soil thickness, 0.25 for confidence, 0.20 for clearance, and 0.10 each for relative height and roughness. Scores are clipped to $[0,1]$. Relative height is normalized by the maximum soil thickness within each connected component. Offline candidate thresholds are soil height $\geq0.15$~m, confidence $\geq0.70$, relative height $[0.15,0.80]$, slope $[10,45]^\circ$, and score $\geq0.45$. Temporal selection uses a three-cell association radius, three-frame confirmation, a 0.10 replacement-score margin, and at most ten-frame dropout retention. The association radius does not cap accepted replacement displacement.

\subsubsection{Motion policy}
\label{sec:motion_setup}
The 38-D observation comprises cylinder positions and velocities (6), swing sine, cosine, and rate (3), tip position and velocity (6), three waypoint-error vectors (9), path progress, tube deviation, and terminal flag (3), task indicator (2), episode progress (1), previous action (4), and current pitch, target pitch, pitch error, and pitch rate (4). An adapter maps base-frame kinematics to training coordinates.

The RL Tracker uses a two-layer, 256-unit MLP with observation normalization and four actions, trained for two million steps. PPO uses batch/buffer sizes 1024/10240, learning rate $10^{-4}$, clipping 0.2, GAE 0.95, discount 0.99, three epochs, and horizon 128. Both task modes use deployment-matched speed limits and action filtering. Target pitch is $70^\circ$ for approach and $180^\circ$ for transport, measured as the signed bucket-axis angle relative to the upper structure's forward axis in its sagittal plane, which rotates with swing.

The progress, distance, path, and pitch weights in Eq.~\eqref{eq:reward} are 5, 0.3, 0.5, and 0.2, respectively. Progress is normalized waypoint-index progress, updated on arrival. Distance and pitch errors are normalized by 1.13~m and $180^\circ$; the path penalty is $c_t=\max(d_t^{\mathrm{path}}/0.04-1,0)$, where $d_t^{\mathrm{path}}$ is the bucket-tip distance to the reference path in meters. Mean squared action change and magnitude have weights 0.05 and 0.01, with a 0.001 step cost. Arrival bonuses are 1 per waypoint and 10 for path completion.

The dead-zone cost has weight 0.03 and uses $D=\min(1.1d_{\mathrm{dz}},1)$ for configured direction-specific threshold $d_{\mathrm{dz}}$. The final-waypoint braking cost is $r_t^{\mathrm{stop}}=\max(1-d_t/0.5,0)\allowbreak\cdot[0.20v_t^2+0.10\max(v_ta_t^{\mathrm{sw}},0)^2]$, with remaining distance $d_t$ in meters and normalized swing velocity and command $v_t,a_t^{\mathrm{sw}}\in[-1,1]$. This cost is zero in other stages.

\subsubsection{Digging policy}
\label{sec:act_setup}
ACT uses 104 accepted teleoperation episodes (13,007 synchronized RGB--proprioception--action frames), split by parent episode into 83 training and 21 validation episodes. Inputs are $640\times480$ RGB and 11-D proprioception: cylinder displacements and velocities (6), joint angles (3), and swing angle and rate (2). The policy predicts 20-action chunks and executes ten commands before querying updated observations, without temporal ensembling. Weights remain frozen during deployment.

\subsection{Evaluation Protocols}
\label{sec:evaluation_protocols}

\subsubsection{Offline target stability}
\label{sec:target_protocol}
Three soil resets each provide initial and manually depressed surfaces, yielding six static sequences of 26.2--26.6~s and 1062 maps. We compare global highest-point selection, Spatial Only (spatial filtering and scoring only), and the Temporal Target Selector; history resets per sequence. Valid maps use machine-frame $X\in[0.3,2.0]$, $Y\in[-0.9,1.4]$, $Z\in[-1.0,-0.5]$~m, confidence $\geq0.70$, and at least three points per cell. Highest-point selection takes the maximum-$Z$ valid cell. All methods share recorded starts, obstacles, and an approach offset of 0.2723~m radially outward and 0.5957~m vertically.

We measure adjacent-valid-frame XY displacement, count jumps above 0.1~m, and report output fractions. Missing outputs are not bridged; frames remain nested within sequences and soil resets. Sixteen frozen-candidate parameter variants additionally use a separate XYZ displacement metric. The static sequences test observation-induced variation.

\subsubsection{Local empty-bucket motion}
\label{sec:local_motion}
The RL Tracker and a damped-least-squares (DLS) kinematic controller each perform five 10-cm motions at $45^\circ$ upward, with matched initial states and initial pitch as the orientation reference. Success requires position error within 2~cm in 30~s, without a separate pitch criterion. Metrics are tracking time, terminal kinematic error, and absolute pitch change. Timing spans the first to last recorded tip states and excludes startup, planning, and cleanup.

\subsubsection{Simulation--real task completion}
\label{sec:transfer_protocol}
Ten paired three-waypoint tasks (P01--P10) use identical policy weights in simulation and on the physical machine. Initial tip position and pitch are geometrically aligned; joint observations need not match. Policy frequency is 20~Hz and simulation timestep 0.01~s. Intermediate and endpoint thresholds are 40 and 25~cm, respectively. We measure completion, tracking duration excluding initialization/cleanup, and 3-D terminal target distance. This task-completion test uses a different tolerance from the 2-cm local-motion comparison.

\subsubsection{Digging-skill payload}
\label{sec:digging_protocol}
ACT is compared with Fixed Dig under a shared sequence of five catalog targets, RL Tracker motion control with terminal braking, and unloading. Fixed Dig uses cylinder-position feedback for entry, curl/pull, and lift. Its normalized targets are $(b,s,q)=(0.26,0.07,0.90)$, $(s,q)=(-0.20,-0.89)$, and $b=-0.30$, respectively, for boom $b$, stick $s$, and bucket $q$. Active axes receive signed 0.5 velocity commands until within 0.07 tolerance; completed and inactive axes receive zero. Each stage has a 0.15-s zero hold and 10-s timeout. ACT retains joystick semantics under the same final actuator limits.

Four paired five-scoop blocks yield 20 weighed completed cycles per method. ACT runs first in pairs 1 and 4, Fixed Dig in pairs 2 and 3. Soil is reset to nominally 0.30~m above the pool floor before each block, without intermediate leveling. ACT's 400,000-step checkpoint terminates at a 130-step cap or an inactive chunk; Fixed Dig uses position-based completion. Thus, payload compares complete skills under their respective stopping rules. Ten development scoops and three interrupted starts (one approach cancellation, two ACT readiness failures) are excluded from completed-cycle analysis.

We report block totals and mean completed-cycle payload, treating paired blocks as the replication unit. An exploratory 95\% percentile interval for mean block difference enumerates all $4^4=256$ paired bootstrap resamples.

\subsubsection{Physical target selection}
\label{sec:physical_target_protocol}
Six paired comparisons use the same ACT policy with the Temporal Target Selector or global highest-point selection: three pairs on leveled soil and three on a central depression. Each pair includes one cycle per method with alternating order. Both preserve selected $X/Y$, add 0.6~m to $Z$, and use the approach orientation.

\subsubsection{Integrated operation and handoffs}
\label{sec:continuous_protocol}
Three completed five-scoop runs assess consecutive operation as the pile changes (Fig.~\ref{fig:continuous_photos}). Execution adds 0.6~m to target $Z$ and applies a 0.15-m pool-edge margin, $X\geq0.9$~m, and $Z\leq-0.2$~m before planning, so executed and raw targets can differ. All runs allow 8~s for selection; S2 and S3 also permit a 12-s retry, which is never triggered. Three-point references use a midpoint that averages endpoint radii and heights about the swing axis and bisects the shortest angular displacement; transport raises it to at least target height. Pre-motion observations fix the ground reference; body-filtered soil points are height-clipped before cell-maximum extraction.

We report run payload, per-scoop payload, duration, interventions, and return status. Scoops and transitions are nested within the three observed run sequences. We also report data from developer teleoperation (20 scoops) and professional demonstrations (104 scoops, also used for ACT). Human times cover approach, digging, transport, unloading, and swing, excluding weighing; autonomous times additionally include startup and finalization.

At controller handoff, command authority is revoked and a zero command confirmed before the next policy takes control. Handoff time spans the outgoing terminal-zero STM32 acknowledgment to the incoming first-nonzero acknowledgment on one Jetson monotonic clock. This metric characterizes command transfer at the acknowledgment boundary.

\section{Results}
\label{sec:results}

Results follow the six protocols in Section~\ref{sec:evaluation_protocols}, from individual components to integrated operation.

\begin{figure}[!t]
\centering
\includegraphics[width=\columnwidth]{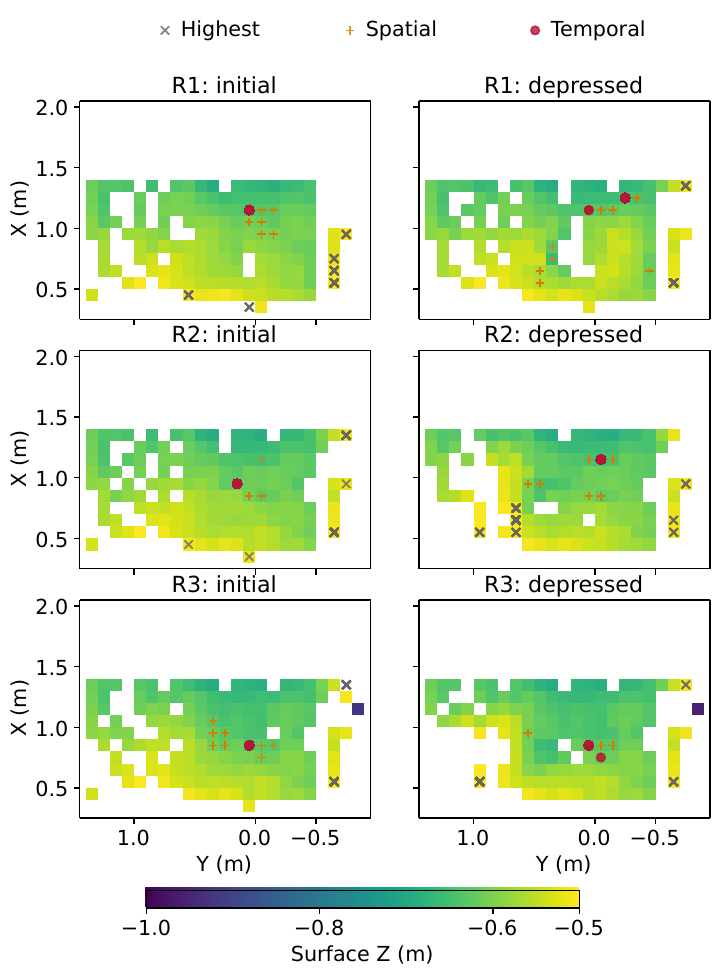}
\caption{Offline target selections over middle-frame elevation maps. Rows show three soil resets; columns show initial and manually depressed surfaces. Highest, Spatial, and Temporal denote highest-point selection, Spatial Only, and the Temporal Target Selector; markers show all valid targets. Shared colors encode machine-frame surface $Z$; blank cells lack valid height.}
\label{fig:e5a_maps}
\end{figure}

\subsection{Offline Target Stability}

\begin{table}[!htb]
\caption{Static-terrain target selection. H: global highest point; S: Spatial Only; F: Temporal Target Selector. Jumps are adjacent-valid-frame XY changes above 0.1~m. I/C denote initial/depressed surfaces within a reset.}
\label{tab:target_results}
\centering
\scriptsize
\setlength{\tabcolsep}{4pt}
\begin{tabular}{@{}lrrrrr@{}}
\toprule
Sequence & Frames & H jumps & S jumps & F jumps & F outputs \\
\midrule
R1 I & 161 & 61 & 50 & 0 & 159 \\
R1 C & 155 & 44 & 98 & 0 & 151 \\
R2 I & 176 & 67 & 57 & 0 & 174 \\
R2 C & 204 & 48 & 82 & 0 & 202 \\
R3 I & 186 & 46 & 133 & 0 & 184 \\
R3 C & 180 & 34 & 103 & 1 & 178 \\
\midrule
Total & 1062 & 300 & 523 & 1 & 1048 \\
\bottomrule
\end{tabular}
\end{table}

The Temporal Target Selector produces one XY jump across 1041 adjacent valid-output pairs, versus 300 and 523 across 1056 pairs for highest-point and Spatial Only (Table~\ref{tab:target_results}). It returns 1048/1062 targets (98.68\%); both alternatives return a target in every frame. Missing outputs explain the different pair counts. Spatial Only jumps more often than highest-point selection, while temporal selection sharply reduces switches at the cost of occasional missing outputs.

The Temporal Target Selector concentrates selections on both initial and depressed surfaces (Fig.~\ref{fig:e5a_maps}). Every emitted target passes the common offline planner; the observed gain therefore concerns temporal stability. Across sixteen frozen-candidate parameter variants, output fractions are 96.13--99.51\% and sequence-mean adjacent XYZ displacement is 0.00032--0.00370~m. These sensitivity results use the separately defined XYZ metric.

\subsection{Empty-Bucket Motion}

Both controllers meet the 2-cm tolerance in all five trials (Table~\ref{tab:motion_results}). The RL Tracker reduces mean tracking time from 1.93 to 1.04~s (46.1\%), with greater variability (SD 0.54 versus 0.11~s). Terminal kinematic errors remain similar, and mean absolute pitch change is smaller. The result supports faster average execution for this local motion.

\begin{table}[!htb]
\caption{Empty-bucket 10-cm motion at $45^\circ$ upward, five trials per controller. Time is mean $\pm$ sample SD; other continuous metrics are means. Position errors use the machine's kinematic estimate.}
\label{tab:motion_results}
\centering
\small
\begin{tabular}{@{}lrr@{}}
\toprule
Metric & RL Tracker & DLS \\
\midrule
Position success & 5/5 & 5/5 \\
Tracking time (s) & $1.04\pm0.54$ & $1.93\pm0.11$ \\
Terminal position error (cm) & 1.62 & 1.77 \\
Absolute pitch change ($^\circ$) & 2.53 & 3.56 \\
\bottomrule
\end{tabular}
\end{table}

\subsection{Simulation--Real Task Completion}

\begin{figure}[t]
\centering
\includegraphics[width=\columnwidth]{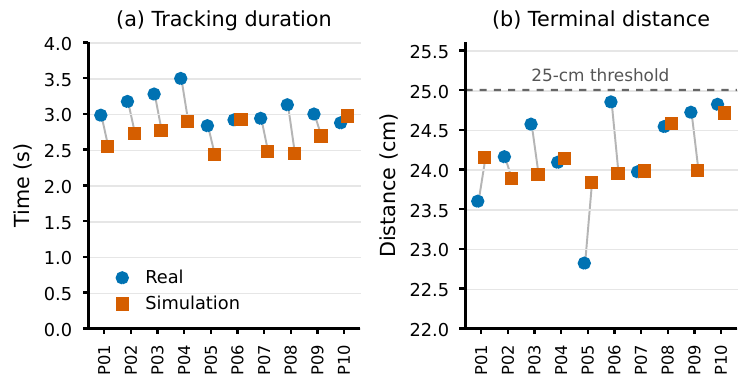}
\caption{Ten simulation--real trial pairs. Lines connect paired measurements; P01--P10 denote trials. The terminal-distance axis is expanded around the 25-cm completion threshold (dashed).}
\label{fig:e4_paired}
\end{figure}

The same policy completes all ten paired tasks in both domains (Fig.~\ref{fig:e4_paired}). Tracking duration averages $3.061\pm0.207$~s on the machine and $2.686\pm0.204$~s in simulation, a mean difference of 0.375~s. Terminal distances are $24.21\pm0.64$ and $24.12\pm0.30$~cm, respectively. Both lie near the 25-cm stopping boundary, consistent with task completion under the specified tolerance after transfer.

\subsection{Digging-Skill Payload}

\begin{figure}[t]
\centering
\includegraphics[width=0.92\columnwidth]{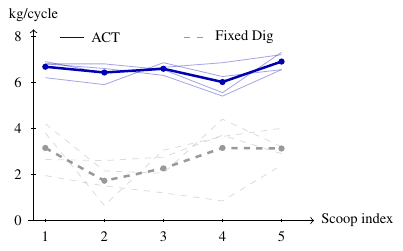}
\caption{ACT versus Fixed Dig completed-cycle payload. Thin lines show four blocks per method; thick lines and points show index-wise means. Each index denotes the same catalog target across blocks, linking target location to execution order.}
\label{fig:payload_curve}
\end{figure}

Across 20 weighed completed cycles per method, ACT delivers 130.40~kg versus 53.65~kg for Fixed Dig, averaging 6.52 versus 2.68~kg per cycle. All four paired blocks favor ACT; block differences span 16.60--25.10~kg, with mean 19.19~kg and exploratory 95\% bootstrap interval [16.83,23.09]~kg. Figure~\ref{fig:payload_curve} shows variation over the shared catalog targets and scoop order. The comparison evaluates the complete deployed skills under the protocol in Section~\ref{sec:digging_protocol}.

\subsection{Physical Target Selection}

\begin{table}[!htb]
\caption{Target-selection comparison: net mass (kg), one cycle per method in each pair. F: leveled soil; D: central depression.}
\label{tab:e5b_results}
\centering
\scriptsize
\setlength{\tabcolsep}{4pt}
\begin{tabular}{@{}llrrr@{}}
\toprule
Pair & Surface & \shortstack{Temporal Target\\Selector} & Highest & Difference \\
\midrule
P1 & F & 6.05 & 6.05 & 0.00 \\
P2 & F & 6.85 & 6.75 & +0.10 \\
P3 & F & 6.55 & 5.50 & +1.05 \\
P4 & D & 6.70 & 6.25 & +0.45 \\
P5 & D & 6.90 & 6.90 & 0.00 \\
P6 & D & 7.05 & 6.70 & +0.35 \\
\midrule
Total & & 40.10 & 38.15 & +1.95 \\
Mean & & 6.68 & 6.36 & +0.325 \\
\bottomrule
\end{tabular}
\end{table}

All twelve cycles complete and return without interruption (Table~\ref{tab:e5b_results}). The Temporal Target Selector yields a mean paired payload difference of 0.325~kg, with four positive pairs and two ties. One pair contributes 1.05~kg of the total 1.95-kg difference. These results complement the offline stability study by demonstrating physical execution with both selectors.

\subsection{Integrated Five-Scoop Operation}

\begin{figure}[!t]
\centering
\includegraphics[width=\columnwidth]{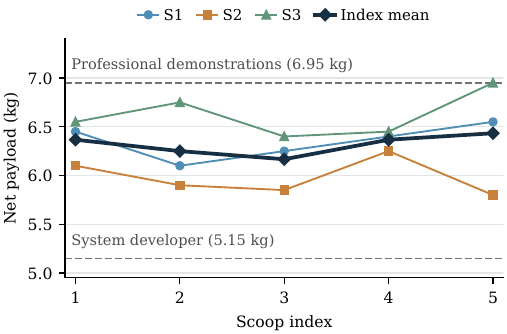}
\caption{Per-scoop payload in three five-scoop autonomous runs. Thin lines show S1--S3; the thick line averages three observations at each index. Dashed lines show means from professional demonstrations (104 scoops, also used for ACT) and teleoperation by the system developer (20 scoops).}
\label{fig:e6_scoops}
\end{figure}

The three analyzed runs each complete five scoops and return without intervention or recorded anomaly (Table~\ref{tab:e6_runs}). They deliver 94.75~kg in total, with mean run payload $31.58\pm1.61$~kg ($n=3$). The fifteen recorded scoop payloads average $6.32\pm0.33$~kg; index-wise means span 6.17--6.43~kg (Fig.~\ref{fig:e6_scoops}), demonstrating continued acquisition through the fifth scoop as the pile changes.

Professional demonstrations and developer teleoperation average 6.95 and 5.15~kg per scoop, respectively. Autonomous durations are 38.8--47.0~s per scoop including startup and finalization, versus 32 and 45~s in the human records. Timing definitions are specified in Section~\ref{sec:continuous_protocol}.

The runs contain thirty RL/ACT handoffs, ten per run and fifteen per direction. RL-to-ACT intervals have median 199.74~ms and maximum 264.00~ms; ACT-to-RL intervals have median 150.42~ms and maximum 164.59~ms. These measurements characterize command transitions at the acknowledgment boundary.

\begin{table}[!htb]
\caption{Five-scoop runs, weighed per run. Duration includes startup and finalization.}
\label{tab:e6_runs}
\centering
\small
\begin{tabular}{@{}lrrr@{}}
\toprule
Run & Scoops & Net mass (kg) & Duration (s) \\
\midrule
S1 & 5 & 31.75 & 204.40 \\
S2 & 5 & 29.90 & 235.14 \\
S3 & 5 & 33.10 & 193.95 \\
\midrule
Total & 15 & 94.75 & --- \\
\bottomrule
\end{tabular}
\end{table}

\section{Discussion}
\label{sec:discussion}

Geometric references let digging locations change without retraining ACT: waypoint following establishes the state from which the demonstration-trained skill acts. This connection depends on bringing the bucket to states where that skill remains applicable. It does not require ACT to receive an explicit terrain goal.

The studies distinguish component benefits from integrated feasibility. Temporal selection reduces observation-induced target jumps; motion trials establish local execution and cross-domain task completion; shared-target trials demonstrate the deployed ACT skill's payload advantage. The five-scoop runs demonstrate the combined capabilities during consecutive operation.

The evaluation covers observation-induced target variation on static terrain, complete digging skills under a shared target sequence, and motion-policy execution. Target location and scoop order are coupled in the digging comparison. Continuous operation is demonstrated over five-scoop sequences on one stationary scaled machine and one soil type.

\section{Conclusion}
\label{sec:conclusion}

We present a framework that connects terrain-aware target selection, task-conditioned RL motion control, and demonstration-trained digging on a scaled hydraulic excavator. Component studies show more stable targets, shorter local motion time, and higher completed-cycle payload against the respective baselines. Three five-scoop runs demonstrate consecutive autonomous operation without human intervention as pile geometry changes. These results show how updated geometric goals can be connected to a reusable local digging skill through waypoint references and coordinated execution on the physical hydraulic platform.

\bibliographystyle{IEEEtran}
\bibliography{references}

\end{document}